\documentclass[review]{elsarticle}
\usepackage{lineno,hyperref}
\modulolinenumbers[5]

\usepackage{times}
\usepackage{helvet}
\usepackage{courier}
\usepackage{url}
\usepackage{stfloats}
\usepackage{floatrow}
\usepackage{graphicx}
\usepackage{epstopdf}
\usepackage{algorithm}
\usepackage{algorithmic}
\usepackage{amsmath}
\usepackage{amssymb}
\usepackage{multirow}
\usepackage{wrapfig}
\usepackage{array}
\usepackage{subfigure}
\usepackage[usenames]{color}
\usepackage{wrapfig}
\usepackage{colortbl,booktabs}
\usepackage{verbatim}
\usepackage{rotating}
\usepackage{setspace}
\usepackage{multirow}
\biboptions{sort&compress}
\begin{document}



\begin{frontmatter}



\title{Knowing Depth Quality In Advance: A Depth Quality Assessment Method For RGB-D Salient Object Detection}


\author{Xuehao Wang$^{1}$
        ~~~~Shuai Li$^{1}$
        ~~~~Chenglizhao Chen$^{2*}$
        ~~~~Aimin Hao$^1$
        ~~~~Hong Qin$^3$
        \\ $^1$State Key Laboratory of VRTS, Beihang University\\
        $^2$College of Computer Science and Technology, Qingdao University\\
        $^3$Stonybrook University}


\begin{abstract}
Previous RGB-D salient object detection (SOD) methods have widely adopted deep learning tools to automatically strike a trade-off between RGB and D (depth), whose key rationale is to take full advantage of their complementary nature, aiming for a much-improved SOD performance than that of using either of them solely.
However, such fully automatic fusions may not always be helpful for the SOD task because the D quality itself usually varies from scene to scene. It may easily lead to a suboptimal fusion result if the D quality is not considered beforehand.
Moreover, as an objective factor, the D quality has long been overlooked by previous work. As a result, it is becoming a clear performance bottleneck.
Thus, we propose a simple yet effective scheme to measure D quality in advance, the key idea of which is to devise a series of features in accordance with the common attributes of high-quality D regions.
To be more concrete, we conduct D quality assessments for each image region, following a multi-scale methodology that includes low-level edge consistency, mid-level regional uncertainty and high-level model variance.
All these components will be computed independently and then be assembled with RGB and D features, applied as implicit indicators, to guide the selective fusion.
Compared with the state-of-the-art fusion schemes, our method can achieve a more reasonable fusion status between RGB and D.
Specifically, the proposed D quality measurement method achieves steady performance improvements for almost 2.0\% in general.
\end{abstract}




\end{frontmatter}




\section{Introduction and Motivation}
Salient object detection (SOD) aims to fast locate the most eye-attractive objects in a given scene~\cite{fang2019visual,CC2015TIP}, and its downstream applications usually include
object detection~\cite{wu2020recent,fu2020oscd},
object segmentation~\cite{hao2020higher,yang2018saliency},
image reconstruction~\cite{li2016multi,CC2019CVPR},
visual tracking~\cite{zhang2020learning,du2020object}
and video saliency detection~\cite{CC2019TIP,CC2019TMM2,CC2017TIP}.
Different to the existing SOD deep models using RGB information solely~\cite{CC2019TMM1,huang2020lightweight,dakhia2019hybrid}, the RGB-D SOD~\cite{cong2017iterative,piao2019depth}, as the main foci of this paper, takes both RGB and D (depth) as input to make a complementary fusion for the SOD task.

In general, RGB-D SOD deep models follow a hypothesis that salient object should be located at a different D layer to the non-salient surroundings nearby~\cite{TCYBSal18}.
Thus, most of the state-of-the-art (SOTA) approaches~\cite{ECCV_P2014} have followed the bi-stream network architecture, the key methodology of which is to compute saliency clues over RGB channels and D channel respectively first and fuse these clues to obtain an RGB-D SOD result later.
Since both RGB and D saliency clues can be easily computed via off-the-shelf deep models that follow the multi-level/multi-scale contrast computations, the fusion procedure is a critical factor for the overall RGB-D SOD performance.

\begin{figure}[!t]
\begin{center}
\includegraphics[width=1\linewidth]{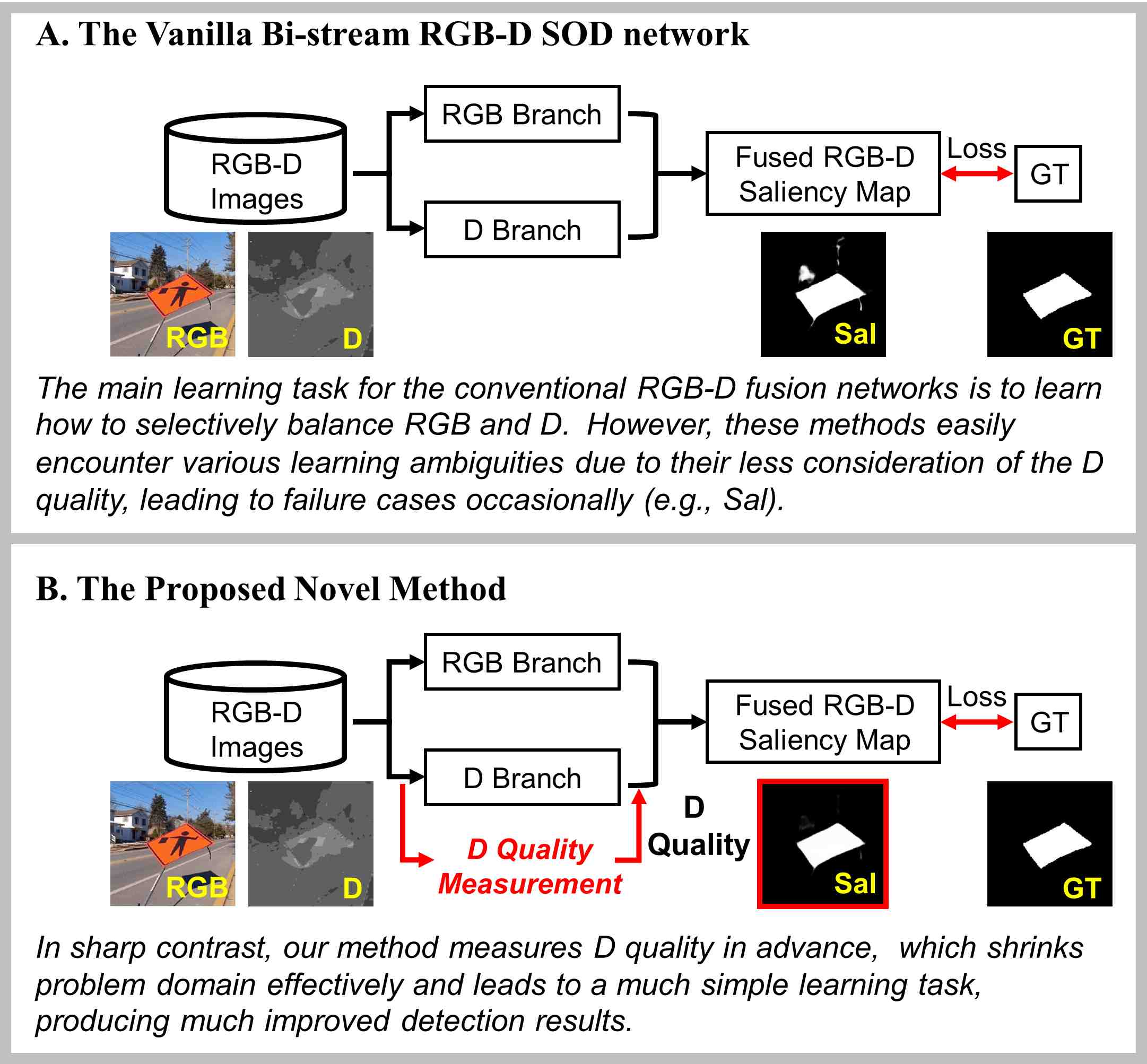}
\end{center}
\vspace{-0.4cm}
\caption{The main difference between the conventional methods and the proposed novel method.}
\label{fig:Integration_guidance}
\end{figure}

In most cases, the widely-used selective fusion (Fig.~\ref{fig:Integration_guidance}-A) is capable of biasing its fused RGB-D saliency map towards either RGB channels or D channel to a certain extent.
By taking numerous RGB-D saliency combinations as training input, the selective fusion models learn how to integrate its two individual inputs, which are respectively the output of RGB branch and D branch, via weighted pixel-wise operations.
Though it can boost the overall performance indeed, the widely-used selective fusion scheme has one critical limitation in common: its unawareness of D quality easily results in various learning ambiguities, leading to a performance bottleneck eventually.
For example, in the face of training instances (i.e., RGB-D images) with high-quality D, the learning scope of the selective fusion scheme should be focused on achieving an optimal combination of RGB and D, which is quite simple and easy in general; however, this learning task will become extremely complex and difficult when the training set contains RGB-D images with various D qualities, which usually correlates to a large problem domain with various learning ambiguities.

To conquer this limitation, we propose a simple yet effective scheme to measure D quality in advance (Fig.~\ref{fig:Integration_guidance}-B), the key idea of which is to devise a series of features in accordance with the common attributes of high-quality D image regions.
To be more concrete, image regions with high-quality D usually have the following attributes:\\
1) Image regions with high-quality D should be capable of separating salient objects from their non-salient surroundings nearby; based on this fact, we resort the edge consistency between RGB and D to measure D quality from a ``low-level'' perspective, which will be further detailed in Sec.~\ref{sec:DQA}.\\
2) In addition, the object-wise homogeneity in D values can constraint the objects' inner regions to be assigned with similar saliency values, which is critical for a complete SOD in face of a salient object that exhibits significant differences in its partial appearances; thus, we propose the regional-wise uncertainty to measure D quality in a ``mid-level'' way, which will be further explained in Sec.~\ref{sec:RGBP}.\\
3) Most importantly, the D quality can be measured from the deep model itself implicitly, i.e., the fused RGB-D saliency can only get improved by using high-quality D, while low-quality D may degenerate the fused saliency; therefore, we shall conduct a ``high-level'' measurement, i.e., computing D quality via the performance variance between deep models that are respectively fed by the 3-dimensional RGB and the 4-dimensional RGB-D, which will be introduced in Sec.~\ref{sec:SALP}.

All these D quality features will be computed independently, and then be assembled to guide selective fusion between RGB and D.
The salient contributions of this paper can be summarized as:
\begin{itemize}
\item
As the first attempt, we have provided a deep insight into the D quality, which is a critical factor for the RGB-D SOD fusion performance, while it has long been overlooked by previous work.

\item
We have proposed a ``multi-level'' D quality measurement to adaptively guide RGB-D saliency fusion, which can effectively alleviate the learning ambiguities and achieve a much-improved SOD performance.

\item
We have conducted extensive quantitative evaluations to prove the effectiveness of the proposed method; we have conducted massive quantitative comparisons to show its performance superiority.

\item
Both source code and results are publicly available at \url{https://github.com/XueHaoWang-Beijing/DQSF}, which may potentially be able to benefit the RGB-D SOD community in the future.
\end{itemize}

\section{Related Work}
The SOTA RGB-D SOD methods usually treat D as an additional image channel, the key rationale of which is to combine RGB saliency and D saliency simply, aiming for the improved overall performance.
Following the vanilla bi-stream fusion methodology, Desingh et al.~\cite{BMVC_K2013} compute the low-level saliency clues over RGB and D channels respectively, then combine these clues to obtain the RGB-D saliency.
Similarly, Ren et al.~\cite{CVPRW_J2015} utilize the multiplicative based fusion to integrate three whole-map saliency features, including D saliency clues, RGB saliency clues and global appearance priors.
Inspired by the phenomenon that salient objects are more likely to be located in front of the image backgrounds, Feng et al.~\cite{CVPR_F2016} propose the depth orientation to measure D saliency.
However, this method easily produces failure detections when the salient objects are not in front of the non-salient backgrounds.

\begin{figure*}[!t]
\begin{center}
\includegraphics[width=1\linewidth]{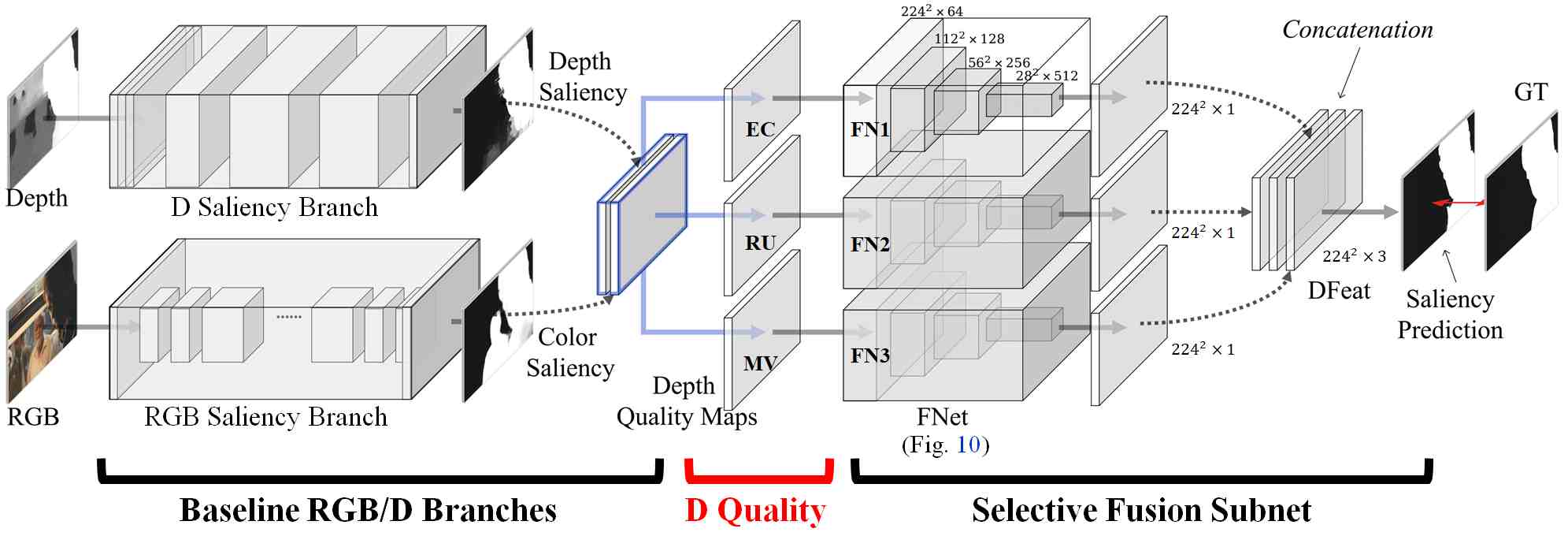}
\end{center}
\vspace{-0.4cm}
\caption{The overall network architecture of our method, where the ``D Quality'' is the main contribution of this paper.}
\label{fig:Network}
\end{figure*}

With the rapid development of deep learning tools, the deep fusion-based SOTA methods are capable of biasing their fusion towards either RGB or D to a certain extent.
Qu et al.~\cite{TIP_Q2017} leverage the convolutional neural networks (CNNs) to selectively fuse multiple low-level handcrafted saliency clues.
Shigematsu et al.~\cite{ICCV_S2017} extract multiple mid-level handcrafted features from depth channel to make the saliency fusion more robust.
Zhu et al.~\cite{Zhu2018PDNet} propose a depth-enhanced network, which consists of two subnetworks; i.e., one master network aims for RGB saliency computation, and the other makes full use of D saliency by integrating its deep features into the master network.
Liu et al.~\cite{liu2019salient} feed the concatenation of original depth channel and RGB channels into a single-stream recurrent convolution neural network based on the multi-scale and multi-level fusion.
Chen et al.~\cite{chen2019multi} conduct RGB-D saliency fusion via a newly designed multi-modal fusion network, which is capable of using multi-scale, multi-path and cross-modal interactions to promote RGB-D SOD performance.
Cong et al.~\cite{Cong2017Saliency} measure the channel-wise importance in advance, and then use it to determine whether the RGB channels or the D channel should be biased during the fusion process.
Liu et al.~\cite{liu2020cross} add the color-stream features into the decoder network of depth stream to overcome the shortcomings of poor-quality depth images and then fuse the multi-modal results under the control of an adaptive gated fusion module.
Zhao et al.~\cite{zhao2019contrast} have mentioned the importance of D quality and introduced a novel RGB based contrast loss into the D stream, aiming to enhance the quality of D features.

\section{Method Overview}
We show the method overview in Fig.~\ref{fig:Network}, which mainly consists of three components: 1) RGB/D baseline branches; 2) D quality; 3) Selective fusion subnet.
Following the vanilla bi-stream structure, the first component includes two individual subbranches, i.e., one for the RGB saliency computation and the other for the D saliency computation, which can be any off-the-shelf deep model.
Next, we resort three individual D quality feature maps, which are measured off-line, to guide the selective fusion between RGB branch and D branch (a.k.a. RGB saliency and D saliency).
This component is the main foci of our paper, each subpart of which will be respectively detailed in Sec.~\ref{sec:RDSS}.
At last, we devise a selective fusion procedure, which is designed with three parallel UNet subbranches, to take full advantage of D quality feature maps, and avoid learning ambiguities.

\section{Depth Quality Measurement}
\label{sec:RDSS}
This section will investigate an effective scheme to conduct D quality assessments from a multi-scale perspective, which includes: 1) low-level edge consistency, 2) mid-level regional uncertainty and 3) high-level model variance.

\begin{figure*}[!t]
\begin{center}
\includegraphics[width=1\linewidth]{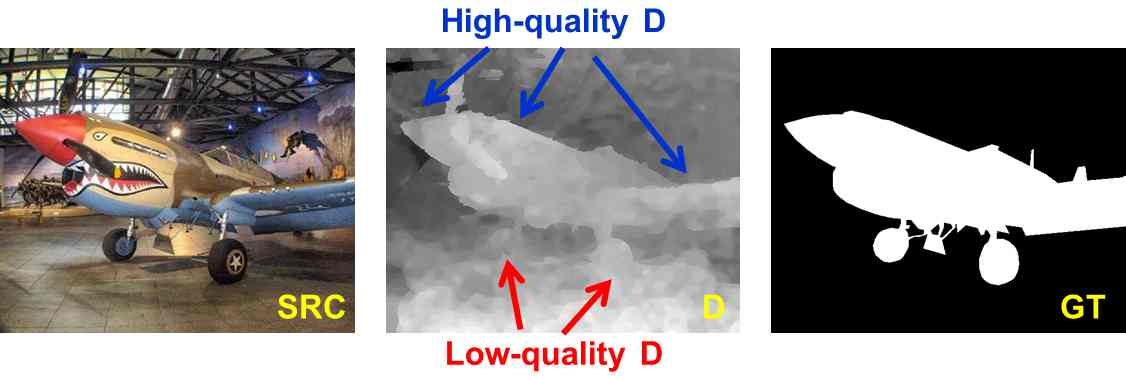}
\end{center}
\vspace{-0.4cm}
\caption{The demonstrations of image regions with different D qualities.}
\label{fig:DDemo}
\end{figure*}

\subsection{Low-level Edge Consistency}
\label{sec:DQA}
\begin{figure}[!t]
\begin{center}
\includegraphics[width=1\linewidth]{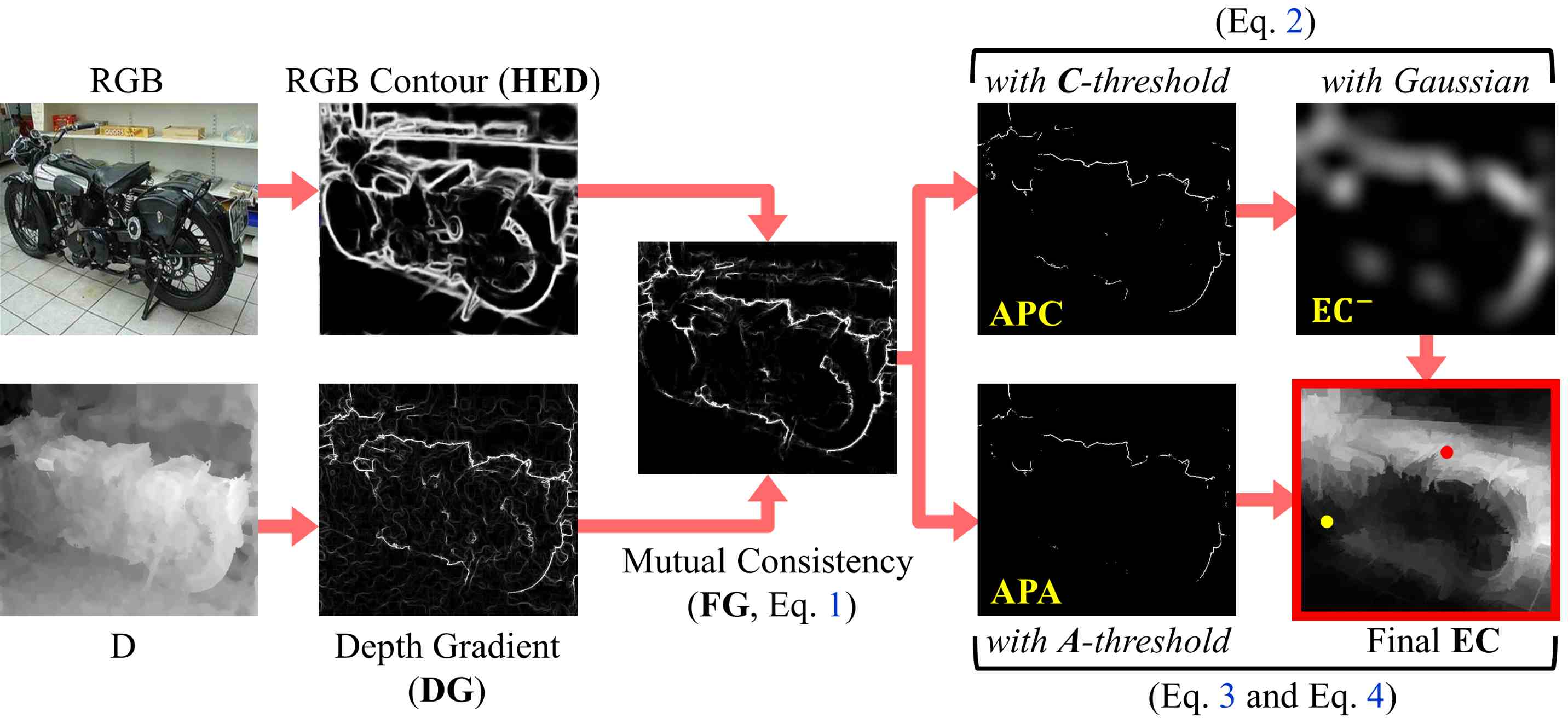}
\end{center}
\vspace{-0.4cm}
\caption{1/3 pipeline of the D quality measurement: low-level edge consistency.}
\label{fig:F1Pipeline}
\end{figure}

Generally, there exists a significant common attribute of the image regions with high-quality D; i.e., it can easily separate salient objects from their non-salient surroundings nearby.
For example, as shown in Fig.~\ref{fig:DDemo}, some parts of the salient object can be easily separated from the non-salient regions by using D channel solely (e.g., the blue arrows), while some parts of the salient object may not be separated easily via D channel (e.g., the red arrows).
In fact, such high-quality D regions are usually located around edges.
In other words, those image regions near the edges that exhibit strong mutual consistency between RGB channels and D channel will have a large potential to be high-quality D regions.
Inspired by this fact, we propose a simple yet effective scheme to locate image regions with high-quality D by measuring the low-level consistency between ``RGB Contour'' and ``D Gradient'' (\textbf{DG}), the method pipeline is demonstrated in Fig.~\ref{fig:F1Pipeline}.

Given a pair of RGB-D images ($\textbf{I}\in\mathbb{R}^{W\times H}$, where $W$ and $H$ respectively represent the width and height), we use the off-the-shelf holistically-nested edge detection method (\textbf{HED}~\cite{ICCV_S2015}) to obtain RGB contour maps.
Compared with the conventional edge detection methods (e.g., Canny), the contour maps produced by \textbf{HED} can highlight object contours while suppressing those less relevant edges located in inner regions of the object.

The mutual consistency (\textbf{FG}) between RGB contour map (\textbf{HED}) and D gradient map (\textbf{DG}) can be simply formulated as Eq.~\ref{eq:MFusion}, thus these pixels with high consistency degree will be interactively compressed.
\begin{equation}
\label{eq:MFusion}
\textbf{FG} = \textbf{DG}\odot \textbf{HED}.
\end{equation}
Here $\odot$ is the element-wise Hadamard product.
We show the pictorial demonstration regarding the mutual consistency map (\textbf{FG}) in the middle column of Fig.~\ref{fig:F1Pipeline}.

Since the high-quality D regions tend to be located near those pixels with large \textbf{FG} values, we determine a subgroup of ``anchor pixels'' (\textbf{APC} in Fig.~\ref{fig:F1Pipeline}) by using a pre-defined hard-threshold ($\rm{T}_c$), and these anchor pixels will be used to coarsely locate the high-quality D regions via Eq.~\ref{eq:RQInit}.
\begin{equation}
\label{eq:RQInit}
\textbf{EC}^-=G\Big(pos(\textbf{FG}-\rm{T}_c)\Big),
\end{equation}
where $\rm T_c$ is a pre-defined hard-threshold; $G$ denotes the Gaussian smoothing (Gaussian Filtering) operation; $pos$ is a function assigning its negative input into zero.

\begin{figure}[!t]
\begin{center}
\includegraphics[width=1\linewidth]{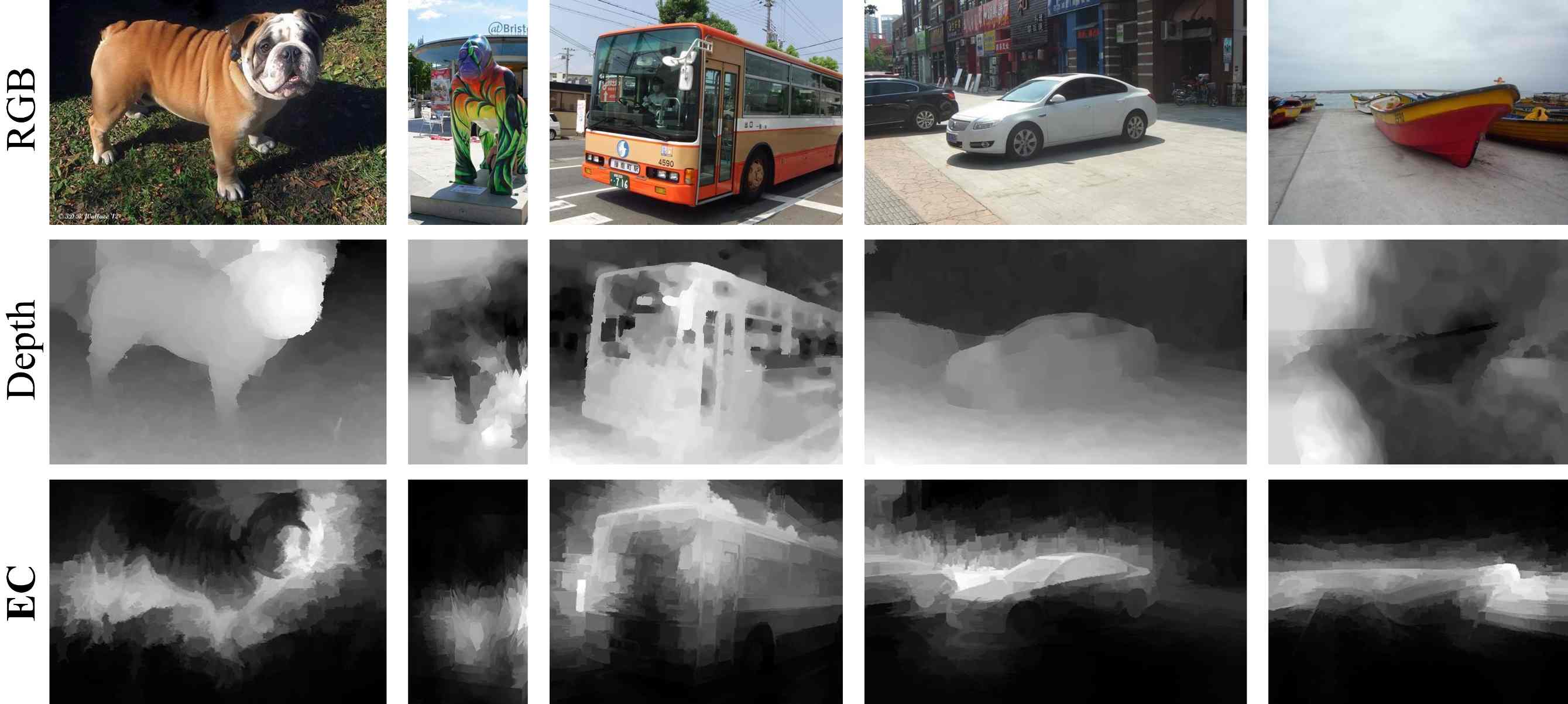}
\end{center}
\vspace{-0.4cm}
\caption{Qualitative demonstrations of the 1/3 D quality map (\textbf{EC}) using low-level edge consistency.}
\label{fig:F1Demo}
\end{figure}

To produce a full regional-wise D quality map, we apply a novel spatial-weighting operation (Eq.~\ref{eq:RQUpdate}) over $\textbf{EC}^-$, which estimates D quality for the image regions that are not quite near edges.
In fact, a typical spatial-weighting scheme should comprise the following two components, i.e., 1) feature similarity measurement (e.g., the $exp$ component in Eq.~\ref{eq:RQUpdate}, we implement it following the common thread mentioned in~\cite{CC2017TIP}); 2) spatial-weighting scope (i.e., the $\phi$ in Eq.~\ref{eq:RQUpdate} that is usually determined by a constant Euclidean distance).
In sharp contrast, the spatial-weighting scope ($\phi$) in our novel method is adaptively determined by a sub-group of most trustworthy anchor pixels via Eq.~\ref{eq:RQUpdate2}, in which these pixels are determined by using an aggressive hard-threshold (i.e., $\rm{T}_a$, and $\rm{T}_a\gg\rm{T}_c$), see \textbf{APA} in Fig.~\ref{fig:F1Pipeline}.
Specifically, we conduct the spatial-weighting over super-pixels (SLIC~\cite{achanta2012slic}) to alleviate the computational burden.
\begin{equation}
\label{eq:RQUpdate}
\textbf{EC}(sp_i) \gets \frac{\sum\limits_{sp_j\in\phi} \textbf{EC}^-(sp_j)\cdot exp\bigg({-\omega_1\cdot\big|\big|c(sp_i),c(sp_j)\big|\big|_2}\bigg)}{\sum\limits_{sp_j\in\phi} exp\bigg({-\omega_1\cdot\big|\big|c(sp_i),c(sp_j)\big|\big|_2}\bigg)},
\end{equation}
\begin{equation}
\label{eq:RQUpdate2}
\phi: \Big|\Big|p(sp_i),p(sp_j)\Big|\Big|_2 \leq min\Bigg\{\Big|\Big|sp_i,pos(\textbf{FG}-\rm{T}_a)\Big|\Big|_2\Bigg\},
\end{equation}
where $sp_i$ denotes the $i$-th superpixel; $\rm{T}_a$ is a predefined aggressive hard-threshold; $\omega_1$ is a strength parameter; function $c(\cdot)$ and $p(\cdot)$ respectively return mean value and center position of their given RGB input.
We demonstrate the final edge consistency map (\textbf{EC}) in the bottom-right of Fig.~\ref{fig:F1Pipeline} (marked with a red rectangle), in which we use a yellow/red dot to indicate some representative low-quality/high-quality D regions. More qualitative demonstrations of \textbf{EC} can be found in Fig.~\ref{fig:F1Demo}.

\subsection{Mid-level Regional Uncertainty}
\label{sec:RGBP}
Though the D quality map measured by the aforementioned edge consistency is generally trustworthy, it still has one major limitation.
That is, the edge consistency based D quality measurement will become less trustworthy if the target image regions are not near object contours, e.g., the inner regions of salient objects.
Thus, in this subsection, we will introduce a novel scheme (i.e., regional uncertainty) to complement the previous edge consistency, aiming for a more comprehensive D quality measurement.

\begin{figure}[t]
      \begin{center}
      \includegraphics[width=1\linewidth]{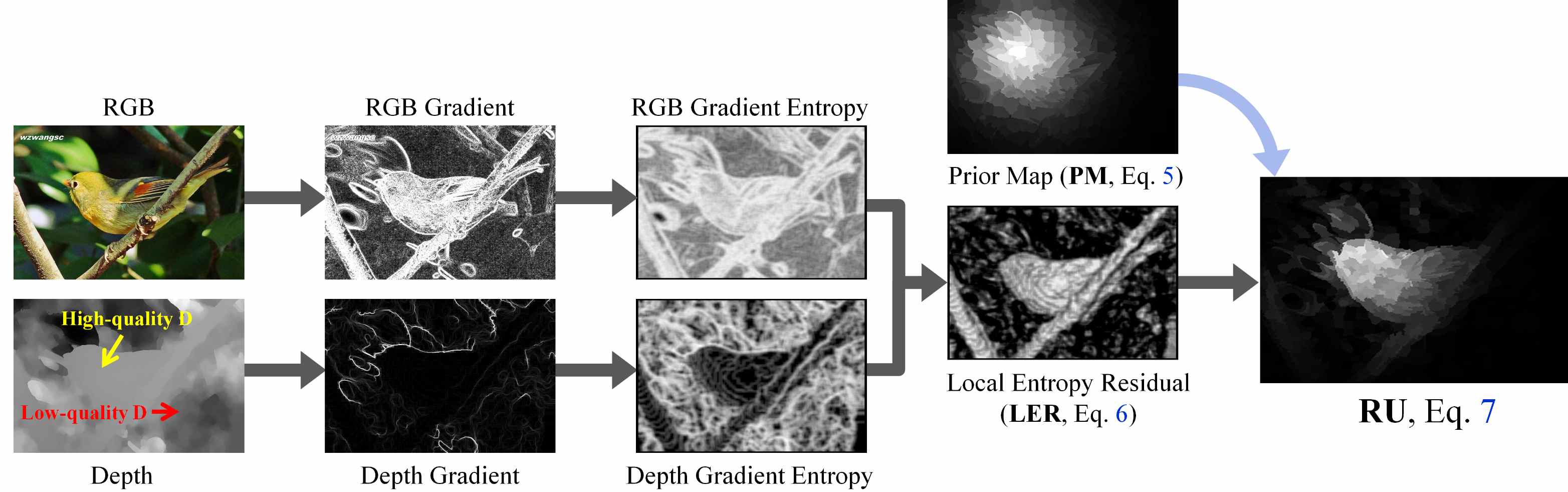}
      \end{center}
      \vspace{-0.4cm}
      \caption{2/3 pipeline of the D quality measurement: mid-level regional uncertainty.}
      \label{fig:F2Pipeline}
\end{figure}

Our regional uncertainty measurement is inspired by another common attribute of the high-quality D regions; i.e., the image regions belonging to an identical object or image area will potentially have high-quality D if their D values tend to be similar with each other.
We show the method pipeline of our regional uncertainty measurement in Fig.~\ref{fig:F2Pipeline}, which mainly comprises two components: 1) localization prior; 2) regional-wise uncertainty.

Though the main interest of this subsection is to conduct D quality assessment for those regions that are not so near to object contours, we should precisely define the concept of ``not so near'' in advance, because it is less trustworthy and not necessary to conduct D quality assessment for the image regions that are far away from salient objects.
Thus, we adopt the localization prior to indicate which regions will be the regions of our interest, i.e., the regions are ``not so near'' to object contours.

We formulate the localization prior ($\textbf{PM}\in\mathbb{R}^{W\times H}$) as Eq.~\ref{eq:PM}.
For example, the localization prior of the $i$-th super-pixel is the summation of L2 spatial distances between it and each of the ``most trustworthy anchor pixels'' that have been introduced in the previous subsection, i.e., \textbf{APA}.
\begin{equation}
\label{eq:PM}
\textbf{PM}(sp_i)= exp\Bigg(-\omega_2\cdot\sum_{l\leq PN}\Big|\Big|p(sp_i),\textbf{HP}_l\Big|\Big|_2\Bigg),
\end{equation}
where $\textbf{PM}(sp_i)$ denotes the localization prior of the $i$-th superpixel; $\textbf{HP}=\xi(\textbf{FG}-\rm{T}_a)\in\mathbb{R}^{PN\times 2}$, denoting the ``most trustworthy anchor pixel'' (i.e., \textbf{APA}), and $PN$ is the total number of \textbf{APA}; \textbf{FG}, $\rm{T}_a$ and $p(\cdot)$ are identical to Eq.~\ref{eq:RQUpdate2}; function $\xi(\cdot)$ returns the coordinates of the non-negative elements; $\omega_2$ is a weighting parameter.
The qualitative demonstration toward $\textbf{PM}$ can be found in the top-right of Fig.~\ref{fig:F2Pipeline} (prior map).

\begin{figure}[!t]
\begin{center}
\includegraphics[width=1\linewidth]{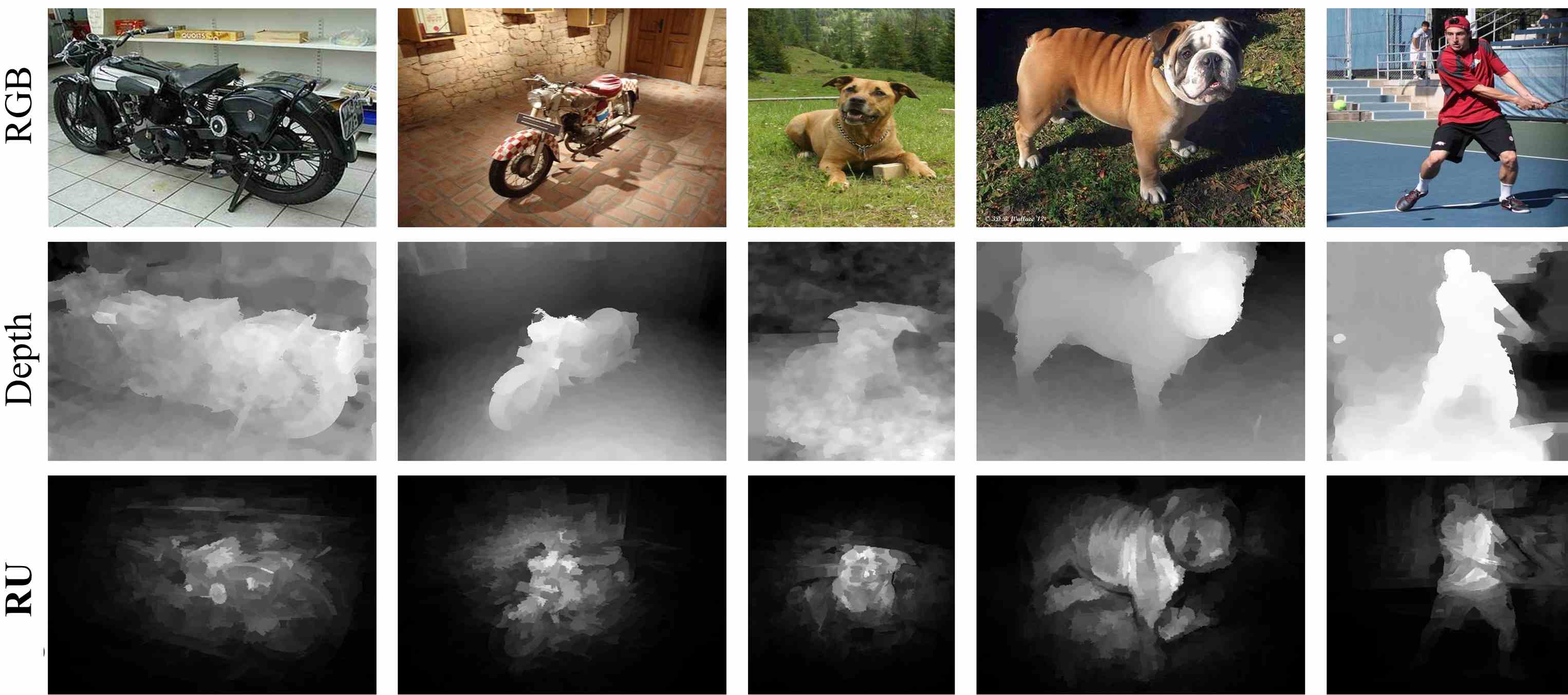}
\end{center}
\vspace{-0.4cm}
\caption{Qualitative demonstrations of the 2/3 D quality map (\textbf{RU}) using mid-level regional uncertainty.}
\label{fig:F2Demo}
\end{figure}

We resort the non-local entropy residual (\textbf{LER}) to represent the regional uncertainty as Eq.~\ref{eq:ER}.
\begin{equation}
\label{eq:ER}
\textbf{LER}(p_i) = pos\Big(E(\textbf{RGBG},\phi_i)-E(\textbf{DG},\phi_i)\Big),
\end{equation}
where \textbf{DG} and \textbf{RGBG} respectively represent D gradient map and RGB gradient map; function $E(\cdot,\cdot)$ returns the entropy value of the image region $\phi_i$.
The qualitative demonstration of \textbf{LER} can be viewed in Fig.~\ref{fig:F2Pipeline}, which usually exhibits large values in the regions with strong homogeneity in D channel yet with a large variance in RGB channels.

We utilize a simple multiplicative based fusion to integrate the previously computed localization prior and local entropy residual (\textbf{LER}) as our mid-level regional uncertainty based D quality map (\textbf{RU}), which can be formulated as Eq.~\ref{eq:SM}).
\begin{equation}
\label{eq:SM}
\textbf{RU} = \zeta\Big(\textbf{LER}\odot\textbf{PM}\Big),
\end{equation}
where $\zeta$ denotes the common thread superpixel-wise spatial-weighting scheme that is identical to the spatial-weighting scheme mentioned in Eq.~\ref{eq:RQUpdate}, the qualitative demonstration of which can be found in Fig.~\ref{fig:F2Demo}.

\begin{figure}[!t]
\begin{center}
\includegraphics[width=1\linewidth]{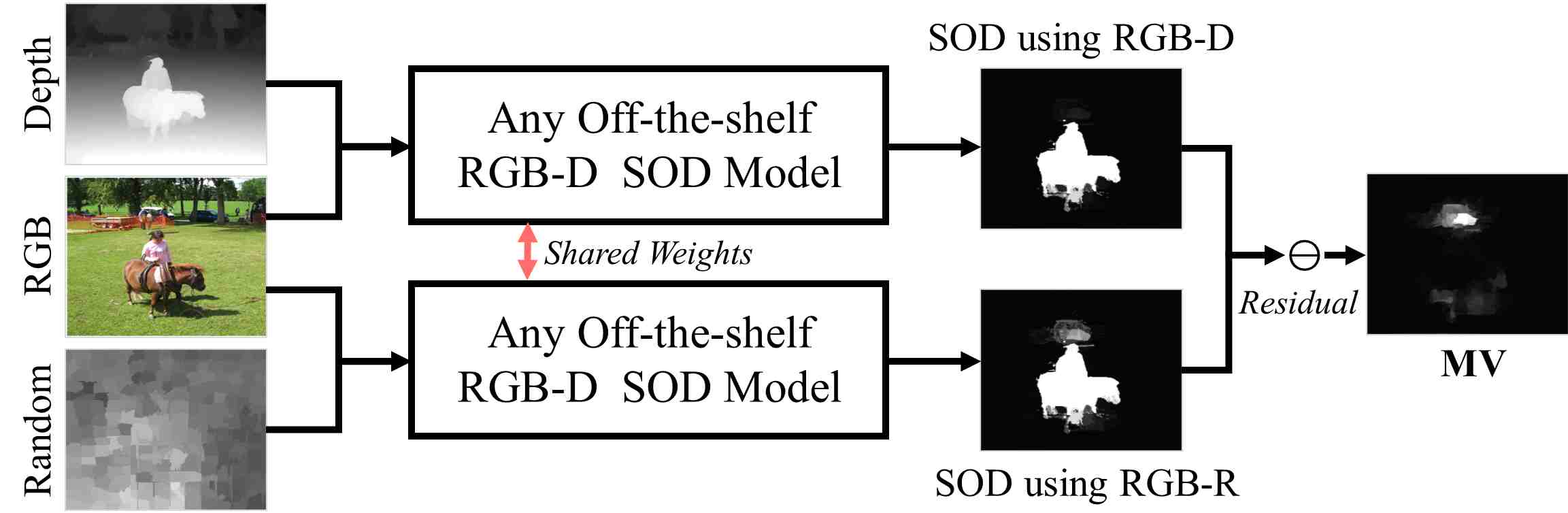}
\end{center}
\vspace{-0.4cm}
\caption{3/3 pipeline of the D quality measurement: high-level model variance.}
\label{fig:F3Pipeline}
\end{figure}

\subsection{High-level Model Variance}
\label{sec:SALP}
In the previous subsections, we have introduced two explicit D quality features (i.e., \textbf{EC} and \textbf{RU}), following a handcrafted methodology.
As another complementary component, in this subsection, we will introduce a novel implicit D quality measurement from the deep model perspective.

This component is inspired by the fact that RGB-D SOD models taking both RGB and D as input can significantly outperform the conventional SOD models using RGB information solely.
Therefore, we propose the high-level model variance as another D quality measurement, the method pipeline can be represented by Fig.~\ref{fig:F3Pipeline}.

\begin{figure}[!t]
\begin{center}
\includegraphics[width=1\linewidth]{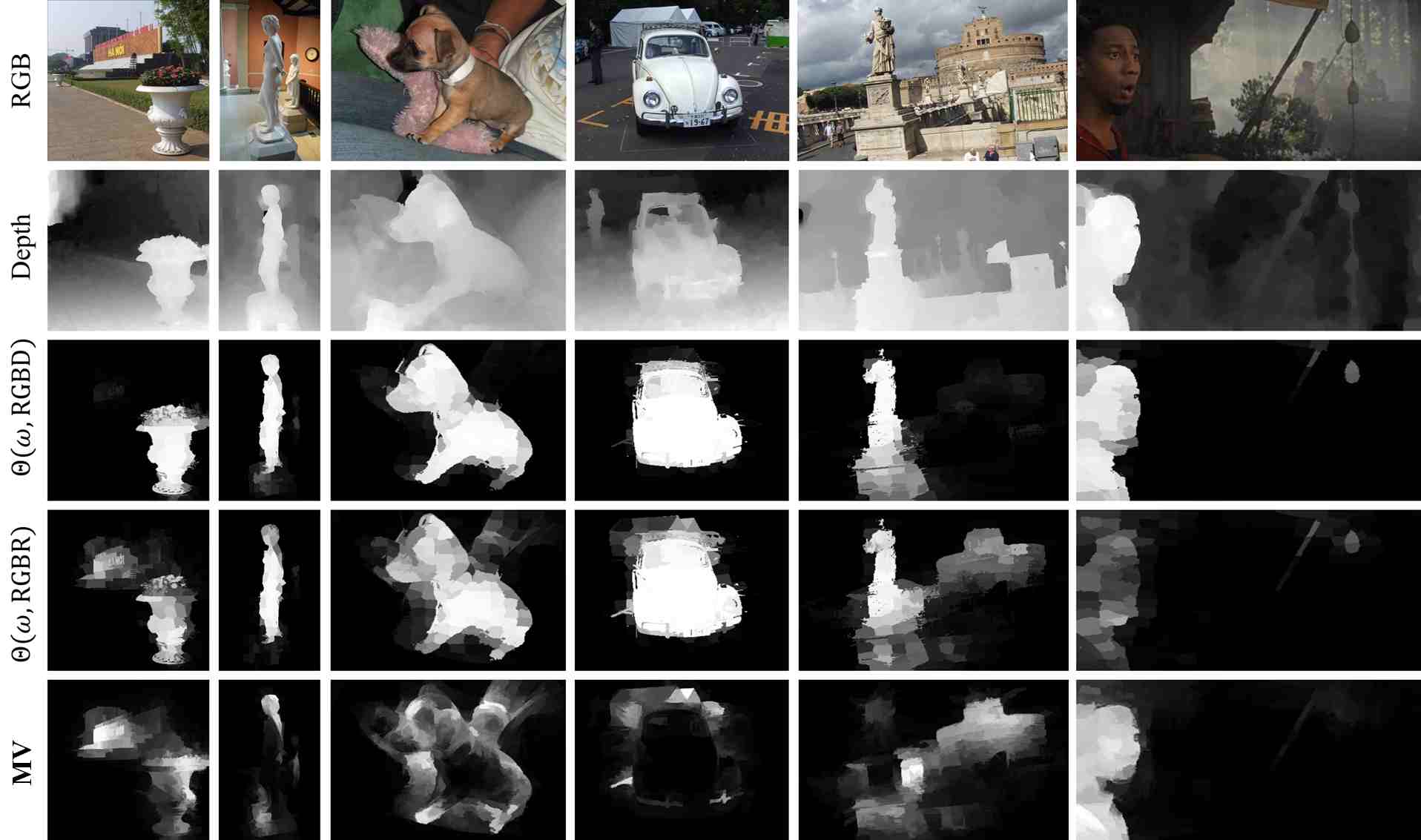}
\end{center}
\vspace{-0.4cm}
\caption{Qualitative demonstrations of the 3/3 D quality map (\textbf{MV}) using high-level model variance.}
\label{fig:F3Demo}
\end{figure}

We use the variance/difference in SOD between two models to measure the implicit D quality, where these two SOD models share identical net architectures and weights yet are fed by different input channels, i.e., RGB-D and RGB-R (the last R represents the ``\emph{Random}'' elements).
We formulate the detailed model variance (\textbf{MV}) as Eq.~\ref{eq:SSM}.
\begin{equation}
\label{eq:SSM}
\textbf{MV} = \Big|\Theta(\omega, \textbf{RGBR})-\Theta(\omega, \textbf{RGBD})\Big|,
\end{equation}
where $|\cdot|$ denotes the obsolete operation; function $\Theta$ represents the pre-trained RGB-D SOD model that receives 4-channel data as input, and $\omega$ denotes the learnable hidden parameters;
\textbf{RGBD} denotes the 4-dimensional RGB-D image and \textbf{RGBR} denotes the newly formulated input data consisting of 3-channel RGB information and 1-channel matrix with random noises.
The \textbf{MV} highlights those D image regions that can benefit the RGB-D SOD task and, consequently, have a large probability to be the high-quality cases.
The qualitative demonstrations of \textbf{MV} can be found in Fig.~\ref{fig:F3Demo}.

\section{Saliency Fusion Guided by D Quality Features}
\label{sec:Network}

\subsection{Fusion Network Overview}
Since the saliency computation over depth channel is theoretically similar to that over RGB channels, our main foci here is to investigate how to make full use of the previously obtained D quality features (i.e., \textbf{EC}, \textbf{RU} and \textbf{MV}) for the RGB-D saliency fusion.

As shown in Fig.~\ref{fig:Network}, our network mainly consists of three components: 1) RGB/D baseline branches; 2) D quality; 3) Selective fusion subnet.
The RGB/D branches can be any off-the-shelf deep models, in which we adopt the off-the-shelf PoolNet~\cite{liu2019simple} as the RGB saliency subbranch, and adopt the pre-trained CPFP~\cite{zhao2019contrast} as the D saliency subbranch, where the CPFP is pre-trained using the same RGB-D training set as our method.

The RGB saliency map and D saliency map will be respectively combined with each of the D quality features.
Thus, the input of the fusion subnet that includes three parallel branches with an identical structure (i.e., FN1, FN2 and FN3) can be represented as $\{RGBSal, DSal, \textbf{EC}\}$, $\{RGBSal, DSal, \textbf{RU}\}$ and $\{RGBSal, DSal, \textbf{MV}\}$.
The fusion subnet (FNet) will be detailed in the next subsection.
It is worth mentioning that a more complex fusion network will lead to better performance, though, we implement it using a lightweight designed structure mainly because this issue is beyond our main interest.

\begin{figure}[!t]
\begin{center}
\includegraphics[width=1\linewidth]{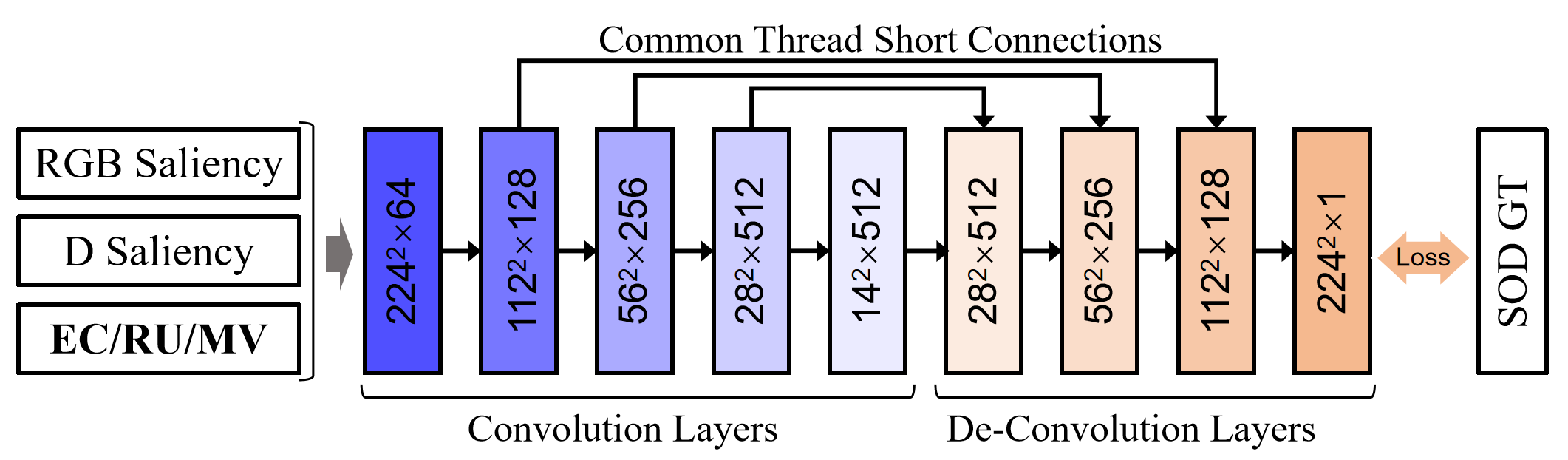}
\end{center}
\vspace{-0.4cm}
\caption{Network architecture of the proposed FNet; the relationship between the FNet and the whole RGB-D SOD net can be found in Fig.~\ref{fig:Network}.}
\label{fig:Baseline}
\end{figure}

\subsection{Fusion Subnet}
\label{sec:fusion}
As shown in Fig.~\ref{fig:Baseline}, each subbranch of the FNet (i.e., FN1, FN2 and FN3, Fig.~\ref{fig:Network}) follows the classic UNet structure to make full use of the multi-scale deep features of the precedent encoder layers, which iteratively integrates the deep features at different encoder layers into each decoder layer.

We use ``DFeat'' to represent the output of FNets (i.e., \textbf{FN1}, \textbf{FN2} and \textbf{FN3}), which can be detailed as Eq.~\ref{eq:DFeat}.
\begin{equation}
\label{eq:DFeat}
\begin{split}
\textbf{DFeat} =  FN1(&\textbf{CSal}, \textbf{DSal}, \textbf{EC})\\
&\cup FN2(\textbf{CSal}, \textbf{DSal}, \textbf{RU})\\
&\ \ \ \ \ \ \ \ \cup FN3(\textbf{CSal}, \textbf{DSal}, \textbf{MV}),
\end{split}
\end{equation}
where the operator $\cup$ denotes the feature concatenate operation, and the obtained \textbf{DFeat} follow the formulation as $\textbf{DFeat}\in\mathbb{R}^{224\times224\times\{1+1+1\}}$.
The final saliency prediction can be obtained by using a convolutional operation (with $1\times1$ kernel) over the \textbf{DFeat}.

\section{Experiments and Evaluations}
\label{sec:Experiments}

\subsection{Datasets}
We have evaluated the proposed method on five public RGB-D benchmark datasets, which are listed below.

\textbf{NJUDS}~\cite{ju2015depth}: This dataset contains 1,985 stereo image pairs, which are gathered from the internet, photographs and stereo movies with optical flow method;
It consists of both simple and complex scenes. \textbf{NLPR}~\cite{ECCV_P2014}: This dataset contains 1,000 images with the depth information captured by Microsoft Kinect in both indoor and outdoor scenes. It is more challenging because its scenes consist of multiple salient objects. \textbf{DES}~\cite{cheng2014depth}: This dataset is also called RGBD135 which contains 135 stereo images captured by Microsoft Kinect in seven indoor scenes. Most of the scenes have a single salient object. \textbf{LFSD}~\cite{CVPR_L2014}: This dataset contains 100 stereo images with the depth information captured by Lytro light field camera. There is no clear boundary between the foreground and background regions in its depth channel. \textbf{STERE}~\cite{niu2012leveraging}: This dataset is also named SSB. It contains 1,000 binocular images captured from both indoor and outdoor scenes.

\subsection{Evaluation Metrics}
We use F-measure~\cite{margolin2014evaluate}, S-measure~\cite{fan2017structure}, E-measure~\cite{fan2018enhanced} and MAE value to evaluate our performance.
The F-measure is related to precision rate and recall rate.
Given a predicted saliency map, we perform binary segmentation with a hard threshold T.
If the obtained foreground is consistent with the ground truth mask, it is deemed as successful detection, and the final precision-recall curves are obtained by varying T from 0 to 255.
As the recall rate is inversely proportional to the precision rate, the tendency of the trade-off between precision and recall can truly indicate the overall detection performance.

F-measure is an important performance indicator when precision rate conflict with recall rate, and it can be computed as Eq.~\ref{eq:FMeasure}, which shows the balance between precision rate and recall rate.

\begin{equation}
\label{eq:FMeasure}
\mathrm{F}\text{-}\mathrm{measure} = \frac{(\beta^2+1)\times\mathrm{PRE}\times\mathrm{REC}}{\beta^2\times\mathrm{PRE}+\mathrm{REC}},
\end{equation}
where $\mathrm{PRE}$ represents the average precision rate,
$\mathrm{REC}$ represents the average recall rate, and
$\beta^2=0.3$ to balance the precision rate and the recall rate.

S-measure is also called Structure-measure~\cite{fan2017structure}.
The novel evaluation focuses on the region-wise and object-wise structural similarities, which is more similar to the human visual system.
It can be formulated as:
\begin{equation}
\label{eq:SMeasure}
\mathrm{S}\textit{-}\mathrm{measure}=\alpha \times S_{o}+(1-\alpha) \times S_{r},
\end{equation}
where we set $\alpha=0.5$ to balance the region-aware ($S_{r}$) and object-aware ($S_{o}$) structural similarity.

E-measure is also named Enhanced-alignment Measure~\cite{fan2018enhanced}.
It combines the pixel-level evaluation (like F-measure) with image-level evaluation (like S-measure) to make a great improvement than other meta-measures.
The formulation of this measure is shown as:
\begin{equation}
\label{eq:EMeasure}
\mathrm{E}\textit{-}\mathrm{measure}=\frac{1}{w \times h} \sum_{x=1}^{w} \sum_{y=1}^{h} \theta_{\rm FM}(x, y)
\end{equation}
where $w$ and $h$ represent the width and height of the image respectively.
$\theta$ is an enhanced alignment matrix~\cite{fan2018enhanced} focused on the pixel-level matching and image-level statistics.
$\rm FM$ represents the foreground map.

The MAE is defined as:
\begin{equation}
\label{eq:MAE}
\mathrm{MAE}=\frac{1}{W \times H} \sum_{x=1}^{W} \sum_{y=1}^{H}|{\rm SM}(x, y)-{\rm GT}(x, y)|,
\end{equation}
where $W$ and $H$ respectively represent the image width and image height;
$\rm SM$ represents the estimated saliency map and $\rm GT$ denotes the ground truth.

\subsection{Implementation Details}
Our training set contains 2050 RGB-D images, including 1400 images from NJUDS dataset and 650 images from NLPR dataset.
All these images are selected the same as~\cite{zhao2019contrast} for a fair comparison. The testing dataset consists of the rest images.
As for our \textbf{MV} feature (Eq.~\ref{eq:SSM}), we train the FNet (Fig.~\ref{fig:Baseline}) for 10K iterations with 4-channel input (RGB-D) over the entire training set.

The parameters we mentioned in Sec.~\ref{sec:RDSS} will be detailed as follows:
we assign the strength parameter $\omega_1$ (Eq.~\ref{eq:RQUpdate}), $\omega_2$ (Eq.~\ref{eq:PM}), the superpixel numbers in Eq.~\ref{eq:RQUpdate} and Eq.~\ref{eq:PM} respectively as \{0.01, 7, 400, 1000\}.
Also, the conservative hard-threshold $\rm T_c$ (Eq.~\ref{eq:RQInit}) and the aggressive hard-threshold $\rm T_a$ (Eq.~\ref{eq:RQUpdate2}) are respectively set as \{20, 30\} times of the average of \textbf{FG} (Eq.~\ref{eq:MFusion}).
Particularly, we use the Gaussian smoothing (Gaussian Filtering) twice (Eq.~\ref{eq:RQInit}) to initialize the regional-wise depth quality, in which the Gaussian parameters were respectively set to \{80,25\} and \{20,20\}.

We optimize the entire network by using Stochastic Gradient Descent (SGD) with a moment 0.9, weight decay 0.005, iter size 10 and learning rate 1e-7.

\begin{figure}[t]
      \begin{center}
      \includegraphics[width=1\linewidth]{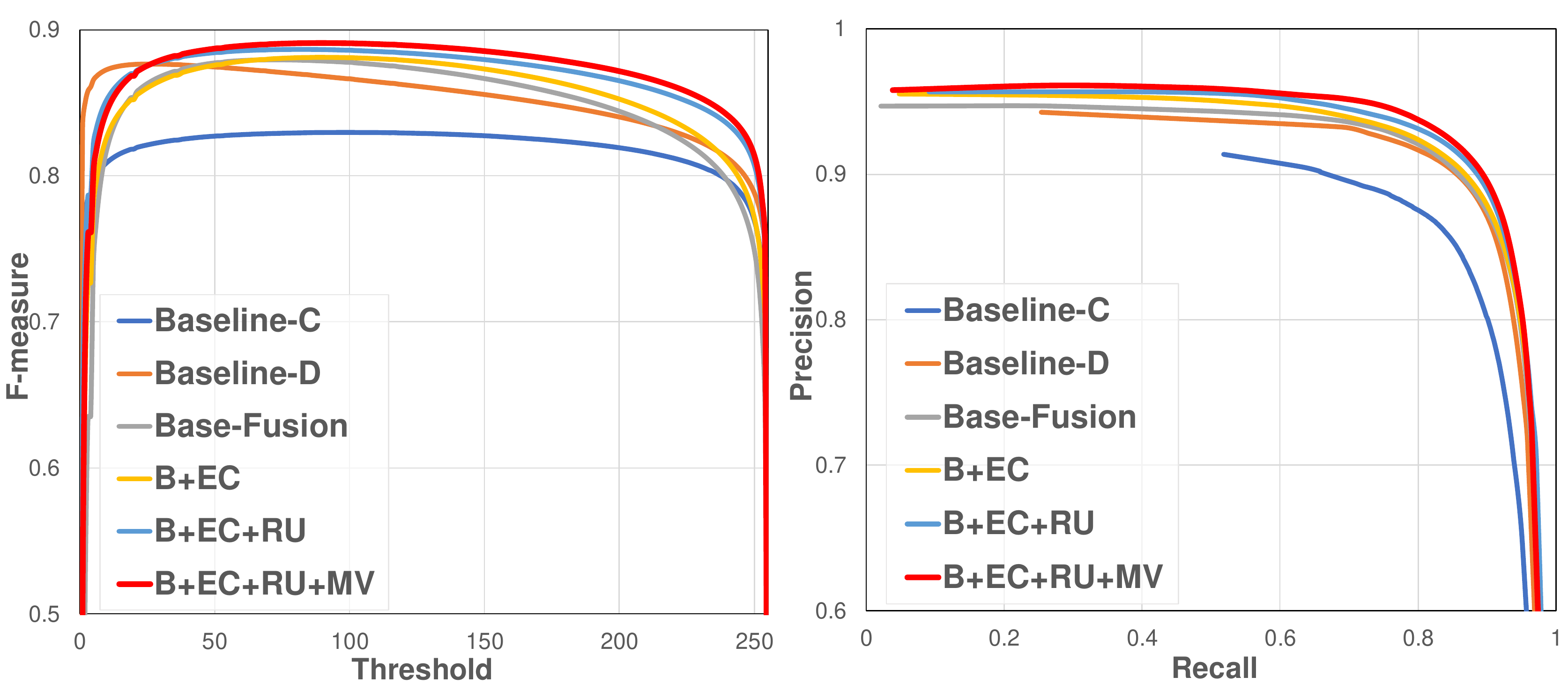}
      \end{center}
      \vspace{-0.4cm}
      \caption{Precision-recall and F-measure curves of different combinations of key components.}
      \label{fig:PR1}
      \end{figure}

\floatsetup[table]{capposition=top}
\begin{table}[t]
  \centering
  \caption{Quantitative evaluations regarding different combinations using various key components.}
  \resizebox{0.8\columnwidth}{!}{
    \begin{tabular}{c|cccc}
    \toprule
          & Sm $\uparrow$& meanF $\uparrow$& maxF $\uparrow$& MAE $\downarrow$\\
    \midrule
    Baseline-C & .843  & .817  & .877  & .077 \\
    Baseline-D & .878  & .850  & .830  & .053 \\
    \midrule
    Base-fusion(B) & .882  & .842  & .879  & .061 \\
    B+EC  & .884  & .851  & .882  & .056 \\
    B+EC+RU & .887  & .863  & .885  & .052 \\
    \midrule
    B+EC+RU+MV & \textbf{.892}  & \textbf{.867}  & \textbf{.891}  & \textbf{.051} \\
    \bottomrule
    \end{tabular}%
  }
  \label{tab:quan_ana}%
\end{table}%

\begin{figure}[t]
      \begin{center}
      \includegraphics[width=1\linewidth]{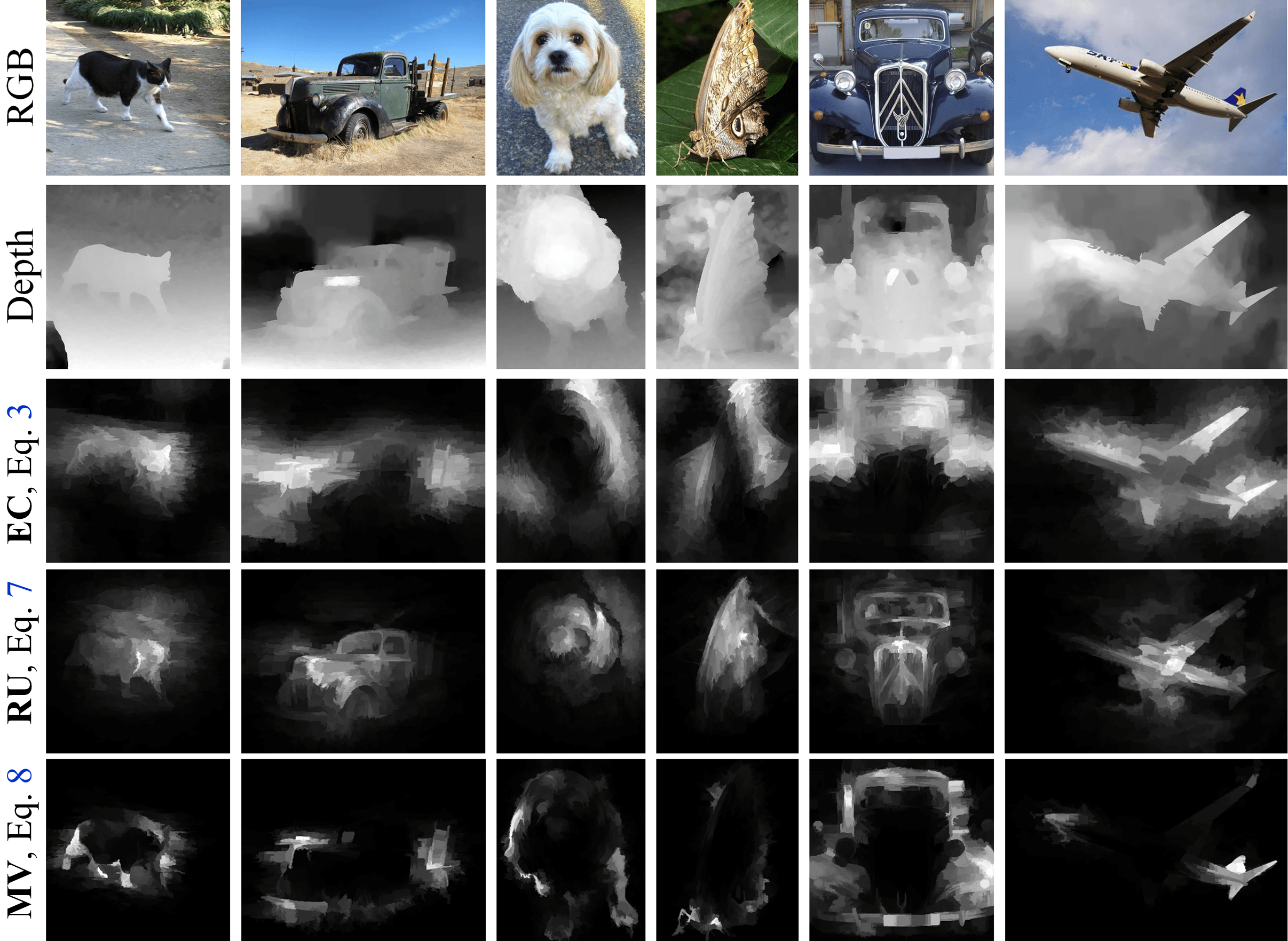}
      \end{center}
      \vspace{-0.4cm}
      \caption{The overall demonstrations of all D quality features, including \textbf{EC}, \textbf{RU} and \textbf{MV}.}
      \label{fig:3F_Demo}
      \end{figure}

\subsection{Component evaluation}
To validate the effectiveness of our method, we perform the component evaluation via S-measure, F-measure and MAE over the NJUDS testing dataset, see details in Table~\ref{tab:quan_ana}.
As one of our baseline sub networks, the ``Baseline-C'' (PoolNet~\cite{liu2019simple}) exhibits the worst performance in Table~\ref{tab:quan_ana}.
Benefited by the usage of depth channel, the ``Baseline-D'' (CPFP~\cite{zhao2019contrast}), which is another baseline subnetwork adopted in our method, exhibits a much-improved detection performance.
The overall performance can be significantly improved by using our selective fusion subnetwork, see the ``Base-fusion'' in Table~\ref{tab:quan_ana}.
Also, we may easily notice a significant performance improvement after integrating the edge-consistency-based D quality feature (\textbf{EC}, Sec.~\ref{sec:DQA}) into the base-fusion network, see the ``B+EC'' in Table~\ref{tab:quan_ana}.
Then, the overall performance can be further improved by further using the regional-uncertainty-based D quality feature (\textbf{RU}, Sec.~\ref{sec:RGBP}) and model-variance-based D quality feature (\textbf{MV}, Sec.~\ref{sec:SALP}), i.e, ``B+EC+RU'' and ``B+EC+RU+MV'', showing the effectiveness of the proposed D quality measurements.
We show PR and F-measure curves of different combinations using various key components in Fig.~\ref{fig:PR1}. We can observe that the model with D quality features (marked as \textbf{B+EC+RU+MV}) achieves the best performance.
The qualitative demonstrations of our D quality feature maps are shown in Fig.~\ref{fig:3F_Demo}, and these feature maps are complementary with each other in general.

\subsection{Performance comparisons to the SOTA methods}
\begin{figure*}[!t]
      \begin{center}
      \includegraphics[width=1\linewidth]{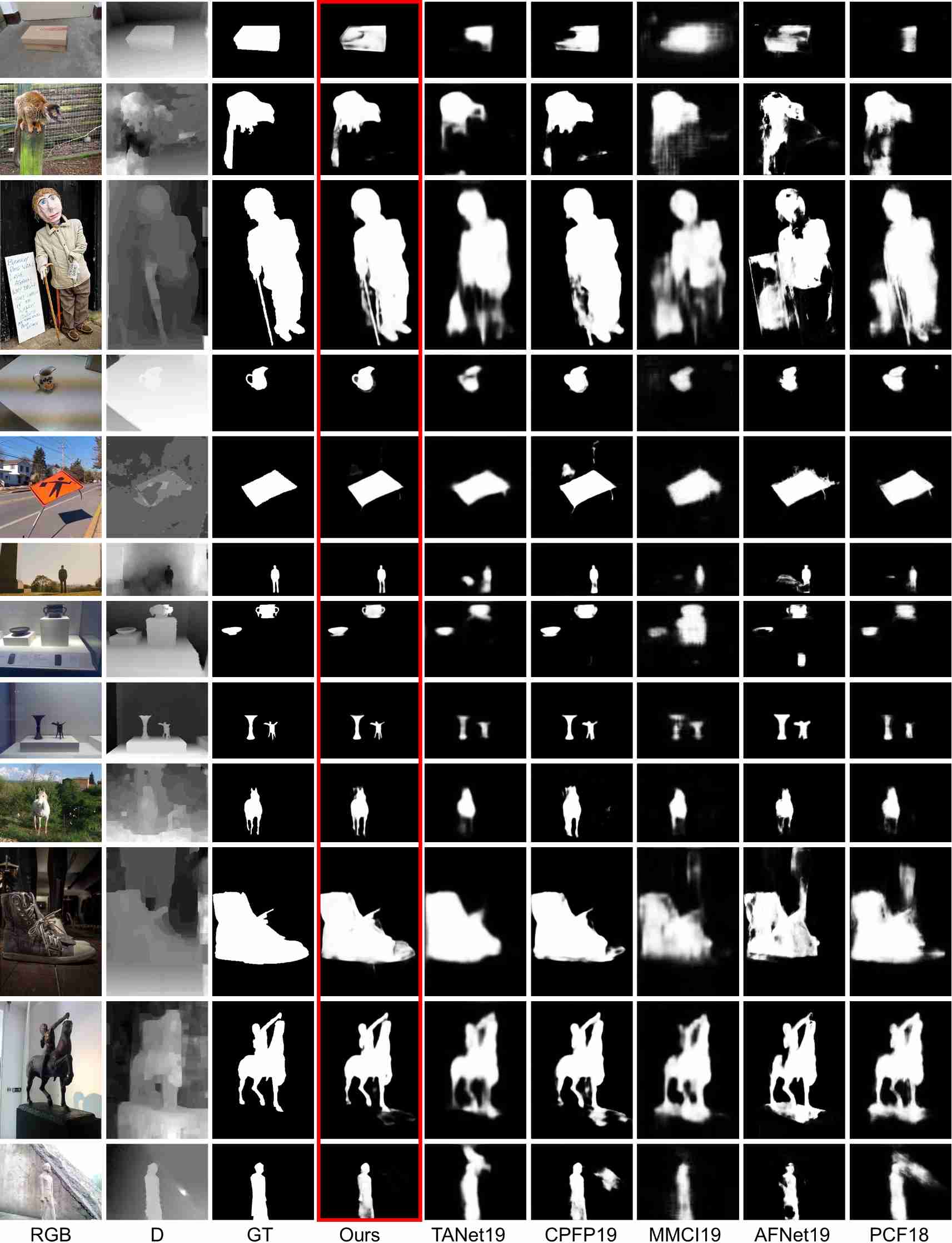}
      \end{center}
      \vspace{-0.4cm}
      \caption{Qualitative comparisons to the SOTA methods. The qualitative comparisons listed here include TANet~\cite{chen2019three}, CPFP~\cite{zhao2019contrast}, MMCI~\cite{chen2019multi}, AFNet~\cite{wang2019adaptive} and PCF~\cite{chen2018progressively}.}
      \label{fig:Demo}
\end{figure*}

In this section, we compare our method with 13 other SOTA approaches, including CPFP~\cite{zhao2019contrast}, TANet~\cite{chen2019three}, MMCI~\cite{chen2019multi}, AFNet~\cite{wang2019adaptive}, PCF~\cite{chen2018progressively}, CTMF~\cite{han2017cnns},
CDB~\cite{liang2018stereoscopic}, DF~\cite{TIP_Q2017}, MDSF~\cite{song2017depth}, CDCP~\cite{zhu2017innovative}, SE~\cite{guo2016salient}, DCMC~\cite{Cong2017Saliency}, LBE~\cite{CVPR_F2016}.
For objective comparisons, all quantitative evaluations are conducted by using the source codes provided by the authors with parameters unchanged.
The detailed quantitative results can be found in Table~\ref{tab:Table_metrix}.
Also, we provide the qualitative comparisons in Fig.~\ref{fig:Demo}, in which our method demonstrates three prominent advantages than these SOTA methods, i.e., 1) more complete detection, 2) rich in saliency details and 3) avoid negative effects induced by low-quality D.
Moreover, for those images with high-quality D, our method can still outperform other SOTA methods.

As shown in Table~\ref{tab:Table_metrix}, the F-measure of our method respectively achieves 1.6\%, 1.6\%, 1.8\%, 2.0\% and 1.6\% improvements over the adopted datasets respectively.
Meanwhile, our method consistently outperforms the SOTA methods in S-measure, E-measure and MAE as well.
We also compare our model with other representative approaches in terms of PR and F-measure curves. As can be seen in Fig.~\ref{fig:PR2} and Fig.~\ref{fig:PR2_2}, our model performs better than all the other approaches.
Specifically, because the proposed edge consistency D quality measurement is developed on the gradient space, the depth-sensing equipment may directly affect the overall performance.
Consequently, we can easily notice that our method performs the best in NLPR dataset (Microsoft Kinect, which can provide high-quality D) and the worst in LFSD dataset (Lytro light field camera, which can only provide low-quality D maps).

\begin{table*}[!t]
      \centering
       \caption{Comparison of quantitative results including F-measure (larger is better), E-measure (larger is better) , S-measure (larger is better) and MAE (smaller is better).} 
      \resizebox{1\columnwidth}{!}{
        \begin{tabular}{c|r|ccccccccccccc|c}
        \toprule
        \multirow{2}[2]{*}{\textbf{Set}} & \multirow{2}[2]{*}{\textbf{Metric}} & LBE & DCMC  & SE & CDCP  & MDSF  & DF    & CDB   & CTMF  & PCF   & AFNet & MMCI  & TANet & CPFP  & \multirow{2}[2]{*}{Ours} \\
              &       & \cite{CVPR_F2016}  & \cite{Cong2017Saliency}  & \cite{guo2016salient}  & \cite{zhu2017innovative}  & \cite{song2017depth}  & \cite{TIP_Q2017}  & \cite{liang2018stereoscopic}  & \cite{han2017cnns}  & \cite{chen2018progressively}  & \cite{wang2019adaptive}  & \cite{chen2019multi}  & \cite{chen2019three}  & \cite{zhao2019contrast}  &  \\
        \hline
        \multirow{7}[2]{*}{\begin{sideways}NJUDS\end{sideways}} & \textbf{Sm $\uparrow$}    & .695  & .686  & .664  & .669  & .748  & .763  & .624  & .849  & .877  & .772  & .858  & .878 & .878 & \textbf{.892} \\
              & \textbf{adpE $\uparrow$}  & .791  & .791  & .772  & .747  & .812  & .835  & .745  & .864  & .896 & .846  & .878  & .893  & .895 & \textbf{.910} \\
              & \textbf{maxE $\uparrow$}  & .803  & .799  & .813  & .741  & .838  & .864  & .742  & .913  & .924 & .853  & .915  & .925 & .923  & \textbf{.928} \\
              & \textbf{adpF $\uparrow$}  & .740  & .717  & .734  & .624  & .757  & .784  & .648  & .788  & .844 & .768  & .812  & .844 & .837  & \textbf{.856} \\
              & \textbf{meanF $\uparrow$} & .606  & .556  & .583  & .595  & .628  & .650  & .482  & .779  & .840  & .764  & .793  & .841 & .850 & \textbf{.867} \\
              & \textbf{maxF $\uparrow$}  & .748  & .715  & .748  & .621  & .775  & .804  & .648  & .845  & .872  & .775  & .852  & .874 & .877 & \textbf{.891} \\
              & \textbf{MAE $\downarrow$}   & .153  & .172  & .169  & .180  & .157  & .141  & .203  & .085  & .059 & .100  & .079  & .060  & .053 & \textbf{.051} \\
        \hline
        \multirow{7}[2]{*}{\begin{sideways}STERE\end{sideways}} & \textbf{Sm $\uparrow$}    & .660  & .731  & .708  & .713  & .728  & .757  & .615  & .848  & .875 & .825  & .873  & .871  & .879 & \textbf{.897} \\
              & \textbf{adpE $\uparrow$}  & .749  & .831  & .825  & .796  & .830  & .838  & .808  & .864  & .897  & .886  & .901  & .906 & .903 & \textbf{.919} \\
              & \textbf{maxE $\uparrow$}  & .787  & .819  & .846  & .786  & .809  & .847  & .823  & .912  & .925  & .887  & .927 & .923  & .925 & \textbf{.932} \\
              & \textbf{adpF $\uparrow$}  & .595  & .742  & .748  & .666  & .744  & .742  & .713  & .771  & .826  & .807  & .829  & .835 & .830 & \textbf{.857} \\
              & \textbf{meanF $\uparrow$} & .501  & .590  & .610  & .638  & .527  & .617  & .489  & .758  & .818  & .806  & .813  & .828 & .841 & \textbf{.861} \\
              & \textbf{maxF $\uparrow$}  & .633  & .740  & .755  & .664  & .719  & .757  & .717  & .831  & .860  & .823  & .863 & .861  & .874 & \textbf{.888} \\
              & \textbf{MAE $\downarrow$}   & .250  & .148  & .143  & .149  & .176  & .141  & .166  & .086  & .064  & .075  & .068  & .060 & .051 & \textbf{.048} \\
        \hline
        \multirow{7}[2]{*}{\begin{sideways}DES\end{sideways}} & \textbf{Sm $\uparrow$}     & .703  & .707  & .741  & .709  & .741  & .752  & .645  & .863 & .842  & .770  & .848  & .858  & .872 & \textbf{.879} \\
              & \textbf{adpE $\uparrow$}  & .911  & .849  & .852  & .816  & .869  & .877  & .868  & .911  & .912  & .874  & .904  & .919 & .927 & \textbf{.944} \\
              & \textbf{maxE $\uparrow$}  & .890  & .773  & .856  & .811  & .851  & .870  & .830  & \textbf{.932}  & .893  & .881  & .928 & .910  & .923 & .931 \\
              & \textbf{adpF $\uparrow$}  & .796 & .702  & .726  & .625  & .744  & .753  & .729  & .778  & .782  & .730  & .762  & .795  & .829 & \textbf{.864} \\
              & \textbf{meanF $\uparrow$} & .576  & .542  & .617  & .585  & .523  & .604  & .502  & .756  & .765  & .713  & .735  & .790 & .824 & \textbf{.831} \\
              & \textbf{maxF $\uparrow$}  & .788  & .666  & .741  & .631  & .746  & .766  & .723  & .844 & .804  & .729  & .822  & .827  & .846 & \textbf{.863} \\
              & \textbf{MAE $\downarrow$}   & .208  & .111  & .090  & .115  & .122  & .093  & .100  & .055  & .049  & .068  & .065  & .046 & .038 & \textbf{.036} \\
        \hline
        \multirow{7}[2]{*}{\begin{sideways}NLPR\end{sideways}} & \textbf{Sm $\uparrow$}    & .762  & .724  & .756  & .727  & .805  & .802  & .629  & .860  & .874  & .799  & .856  & .886 & .888 & \textbf{.900} \\
              & \textbf{adpE $\uparrow$}  & .855  & .786  & .839  & .800  & .812  & .868  & .809  & .869  & .916  & .884  & .872  & .916 & .924 & \textbf{.938} \\
              & \textbf{maxE $\uparrow$}  & .855  & .793  & .847  & .820  & .885  & .880  & .791  & .929  & .925  & .879  & .913  & \textbf{.941} & .932 & .938 \\
              & \textbf{adpF $\uparrow$}  & .736  & .614  & .692  & .608  & .665  & .744  & .613  & .724  & .795  & .747  & .730  & .796 & .823 & \textbf{.858} \\
              & \textbf{meanF $\uparrow$} & .736  & .543  & .624  & .609  & .649  & .664  & .422  & .740  & .802  & .755  & .737  & .819 & .840 & \textbf{.855} \\
              & \textbf{maxF $\uparrow$}  & .745  & .648  & .713  & .645  & .793  & .778  & .618  & .825  & .841  & .771  & .815  & .863 & .867 & \textbf{.884} \\
              & \textbf{MAE $\downarrow$}   & .081  & .117  & .091  & .112  & .095  & .085  & .114  & .056  & .044  & .058  & .059  & .041 & .036 & \textbf{.034} \\
        \hline
        \multirow{7}[2]{*}{\begin{sideways}LFSD\end{sideways}} & \textbf{Sm $\uparrow$}   & .736  & .753  & .698  & .717  & .700  & .791  & .520  & .796  & .794  & .738  & .787  & .801 & .828 & \textbf{.844} \\
              & \textbf{adpE $\uparrow$}  & .770  & .842  & .784  & .780  & .817  & .844  & .703  & .851 & .842  & .810  & .840  & .845  & .867 & \textbf{.883} \\
              & \textbf{maxE $\uparrow$}  & .804  & .856  & .840  & .786  & .826  & .865  & .774  & .865 & .835  & .815  & .839  & .847  & .872 & \textbf{.884} \\
              & \textbf{adpF $\uparrow$}  & .708  & .816 & .778  & .697  & .799  & .806  & .682  & .782  & .792  & .742  & .779  & .794  & .813 & \textbf{.839} \\
              & \textbf{meanF $\uparrow$} & .611  & .655  & .640  & .680  & .521  & .679  & .376  & .756  & .761  & .735  & .722  & .771 & .811 & \textbf{.820} \\
              & \textbf{maxF $\uparrow$}  & .726  & .817 & .791  & .703  & .783  & .817  & .682  & .791  & .779  & .744  & .771  & .796  & .826 & \textbf{.839} \\
              & \textbf{MAE $\downarrow$}   & .208  & .155  & .167  & .167  & .190  & .138  & .218  & .119  & .112  & .133  & .132  & .111 & .088 & \textbf{.086} \\
        \bottomrule
        \end{tabular}}%
      \label{tab:Table_metrix}%
    \end{table*}%

 \begin{figure*}[!h]
      \begin{center}
      \includegraphics[width=1\linewidth]{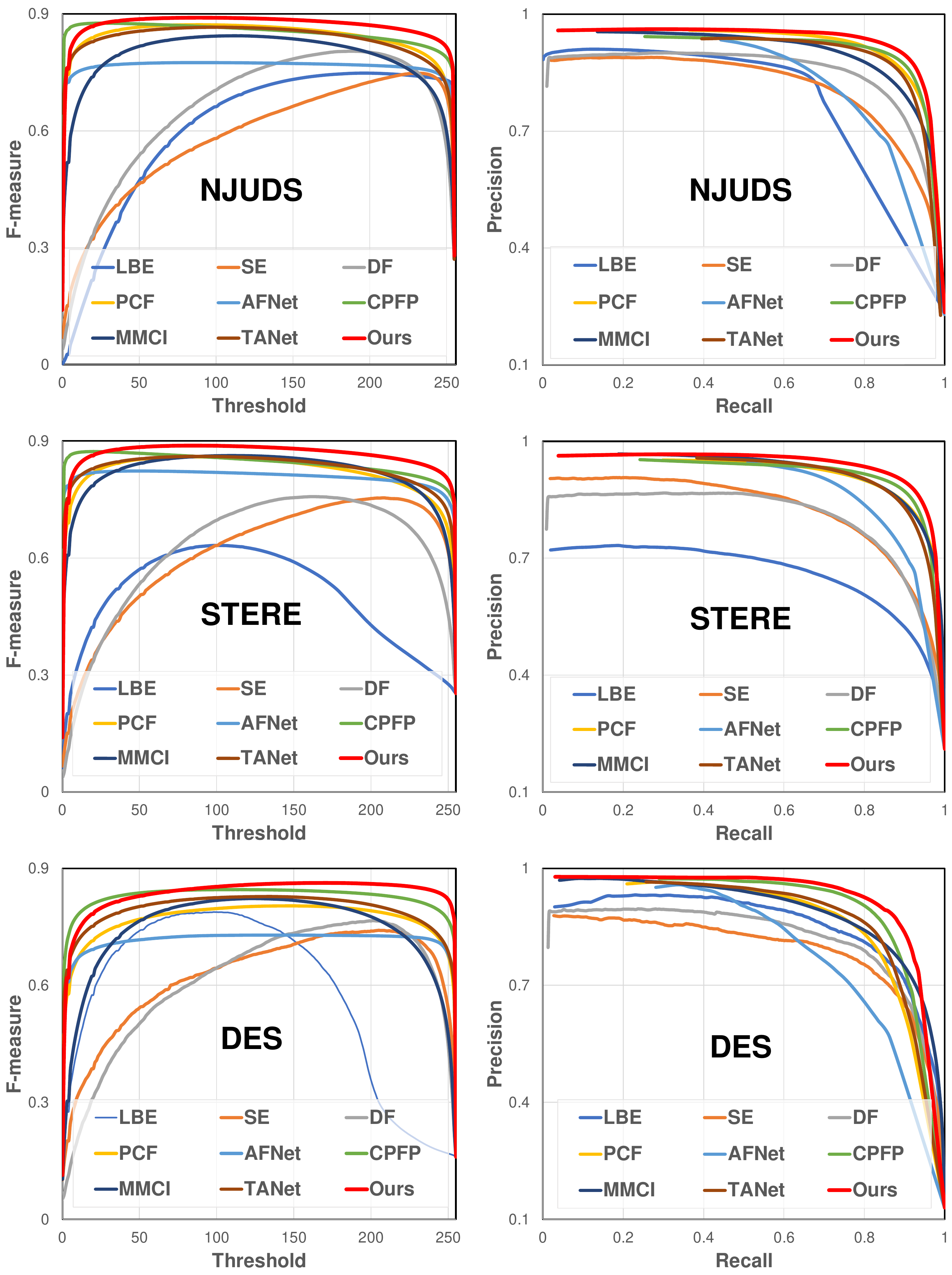}
      \end{center}
      \vspace{-0.4cm}
      \caption{Quantitative comparisons on five popular datasets (3/5) in terms of the PR curves and F-measure curves.}
      \label{fig:PR2}
      \end{figure*}

\begin{figure*}[!h]
      \begin{center}
      \includegraphics[width=1\linewidth]{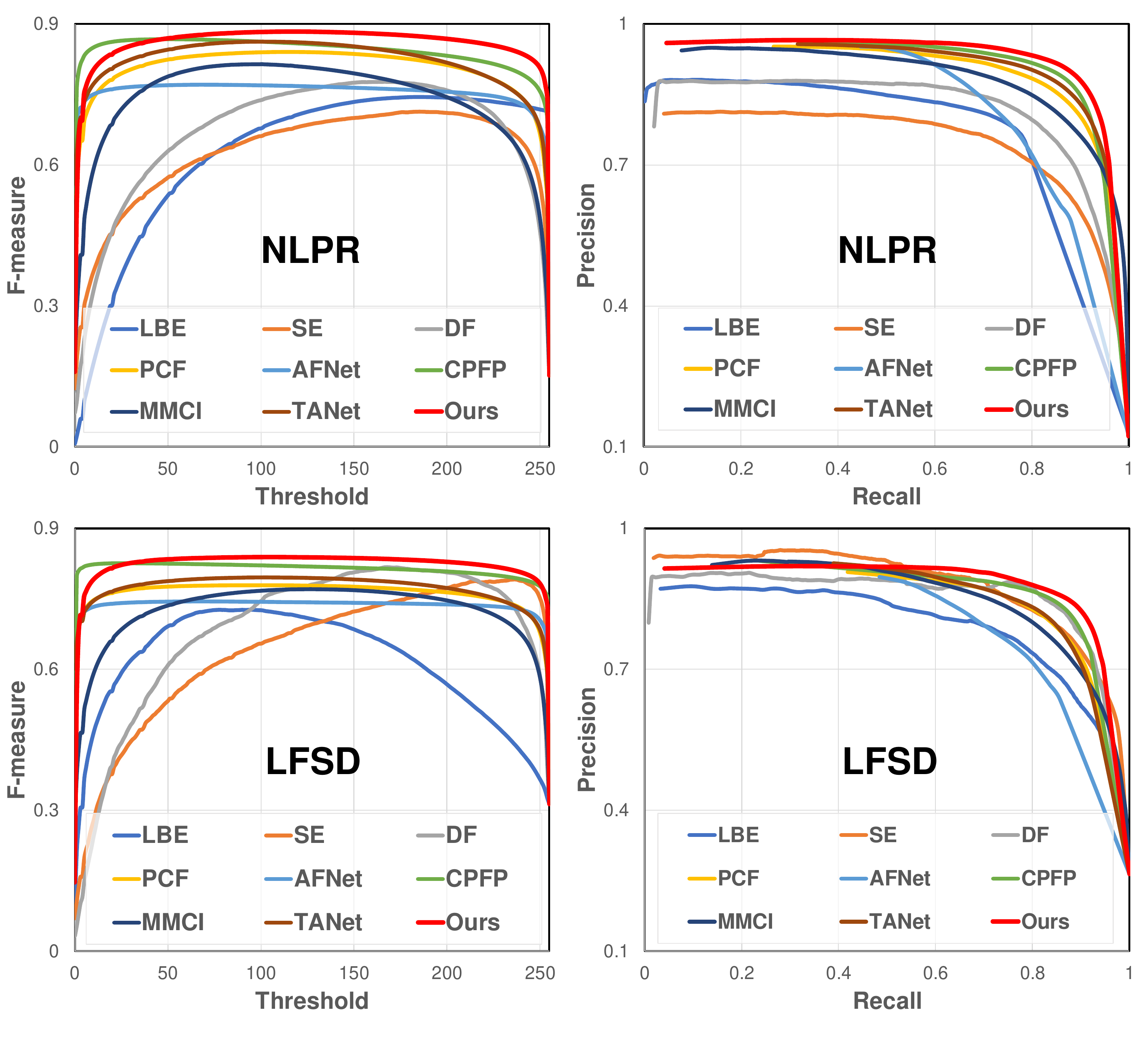}
      \end{center}
      \vspace{-0.4cm}
      \caption{Continued quantitative comparisons on five popular datasets (2/5) in terms of the PR curves and F-measure curves.}
      \label{fig:PR2_2}
      \end{figure*}

\begin{table}[htbp]
      \centering
      \caption{Comparison of model size and test time for each RGB-D image pair with other representative methods.}
      \resizebox{0.8\columnwidth}{!}{
        \begin{tabular}{c|cccc|c}
        \toprule
        Model & PCF~\cite{chen2018progressively} & MMCI~\cite{chen2019multi} & TANet~\cite{chen2019three} & CPFP~\cite{zhao2019contrast} & Ours \\
        \midrule
        Size (MB) & 533.6 & 929.7 & 951.9 & 278.1 & \textbf{235.0} \\
        Time (ms) & ~~65.5 & \textbf{~~51.2}  & ~~70.4  & 170.0 & ~~62.1 \\
        \bottomrule
        \end{tabular}%
      }
      \label{tab:time}%
    \end{table}%

\section{Limitation}
\label{sec:Limitation}
Our fusion network is lightweight designed, whose model size is quite smaller than other representative methods, taking more than 15\% reduction. Its test time also achieves a leading performance.
The quantitative comparisons can be seen in Table~\ref{tab:time}.
However, our method is generally time-consuming due to the handcrafted D quality measurement features.
On a desktop computer with i7-6700 3.40GHz CPU, GTX 1080 GPU and 32GB RAM, it takes almost 0.240s to compute our depth quality-aware features (4.2 FPS with CPU) and another 0.062s (16 FPS with GPU) to make the final saliency prediction.
Thus, our method needs a total of 0.302s to perform SOD for a $224\times 224$ RGB-D image.
Also, our method may produce failure detection if both RGB and D are incapable of separating the salient object from its non-salient surroundings nearby.


\section{Conclusions and Future Work}
\label{sec:Conclusion}
In this paper, we have proposed a novel D quality assessment solution to conduct ``quality-aware'' SOD for RGB-D images.
The SOTA methods easily produce incorrect detections in the face of images with low-quality D.
To conquer it, we have devised three novel features (i.e., \textbf{EC}, \textbf{RU} and \textbf{MV}) to measure the D quality before performing RGB-D saliency fusion.
Meanwhile, we have devised an effective and lightweight designed fusion network to take full use of these D quality features during performing selective RGB-D fusion.
The proposed idea regarding the D quality assessments will have a large potential to benefit the RGB-D SOD community in the future.

As for our near future work, we are particularly interested in developing a novel end-to-end depth quality assessment network, which is capable of measuring the depth quality very fast within a full-automatic manner.
Moreover, we may use D quality maps to complete and enhance D channels.

\vspace{0.3cm}
\centerline{\textbf{\large References}}


\bibliographystyle{IEEEtran}
\bibliography{reference}
\end{document}